\documentclass[electronic]{vgtc}             % electronic version

\ifpdf%                                % if we use pdflatex
  \pdfoutput=1\relax                   % create PDFs from pdfLaTeX
  \usepackage{graphicx}                % allow us to embed graphics files
  \DeclareGraphicsExtensions{.pdf,.png,.jpg,.jpeg} % for pdflatex we expect .pdf, .png, or .jpg files
\else%                                 % else we use pure latex
  \usepackage{graphicx}                % allow us to embed graphics files
  \DeclareGraphicsExtensions{.eps}     % for pure latex we expect eps files
\fi%

\graphicspath{{figures/}{pictures/}{images/}{./}} % where to search for the images

\usepackage{microtype}                 % use micro-typography (slightly more compact, better to read)
\PassOptionsToPackage{warn}{textcomp}  % to address font issues with \textrightarrow
\usepackage{textcomp}                  % use better special symbols
\usepackage{mathptmx}                  % use matching math font
\usepackage{times}                     % we use Times as the main font
\usepackage{cite}                      % needed to automatically sort the references
\usepackage{tabu}                      % only used for the table example
\usepackage{booktabs}                  % only used for the table example
\vgtccategory{Research}

\vgtcinsertpkg

\usepackage{amsmath}
\usepackage{algorithm, algorithmic}

\title{Efficient texture mapping via a non-iterative global texture alignment}

\author{Mohammad Rouhani\thanks{e-mail:mohammad.rouhani@airbus.com. This work was done when Mohammad was with InterDigital.} %
\and Matthieu Fradet\thanks{e-mail:matthieu.fradet@interdigital.com} %
\and Caroline Baillard\thanks{e-mail:caroline.baillard@interdigital.com}}
\affiliation{\scriptsize InterDigital, France}

\abstract{Texture reconstruction techniques generally suffer from the errors in keyframe poses. We present a non-iterative method for seamless texture reconstruction of a given 3D scene. Our method finds the best texture alignment in a single shot using a global optimisation framework. First, we automatically select the best keyframe to texture each face of the mesh. This leads to a decomposition of the mesh into small groups of connected faces associated to a same keyframe. We call such groups fragments. Then, we propose a geometry-aware matching technique between the 3D keypoints extracted around the fragment borders, where the matching zone is controlled by the margin size. These constraints lead to a least squares (LS) model for finding the optimal alignment. Finally, visual seams are further reduced by applying a fast colour correction. In contrast to pixel-wise methods, we find the optimal alignment by solving a sparse system of linear equations, which is very fast and non-iterative. Experimental results demonstrate low computational complexity and outperformance compared to other alignment methods.%
} % end of abstract

\CCScatlist{
 \CCScatTwelve{Computing methodologies}{Computer graphics}{Image manipulation}{Texturing}
\CCScatTwelve{Computing methodologies}{Computer graphics}{Shape modeling}{Mesh models}
\CCScatTwelve{Computing methodologies}{Computer graphics}{Graphics systems and interfaces}{Mixed / augmented reality}
}

\begin{document}


%\firstsection{Introduction}

\maketitle

%% \section{Introduction} %for journal use above \firstsection{..} instead
%This template is for papers of VGTC-sponsored conferences which are \emph{\textbf{not}} published in a special issue of TVCG.

%%__________________________________________________________________________________
\section{Introduction}
\label{sec:intro}

%Intro
Texture reconstruction is a fundamental problem of 3D computer vision with many applications in virtual and augmented reality \cite{Glen18} and interior design \cite{zhang16}. Thanks to commodity sensors and modern visual tracking systems, 3D geometry and texture can be easily reconstructed. However, with the presence of errors in camera poses and illumination changes, it may not result in a high quality reconstruction. In this work we focus on texture reconstruction problem as photometric information is visually more salient than geometry \cite{huang17}. We assume that a 3D mesh is provided along with several keyframes and their associated camera poses.

\begin{figure}%[t]
\centering
\begin{tabular}{cc}
\includegraphics[width = 0.45\columnwidth]{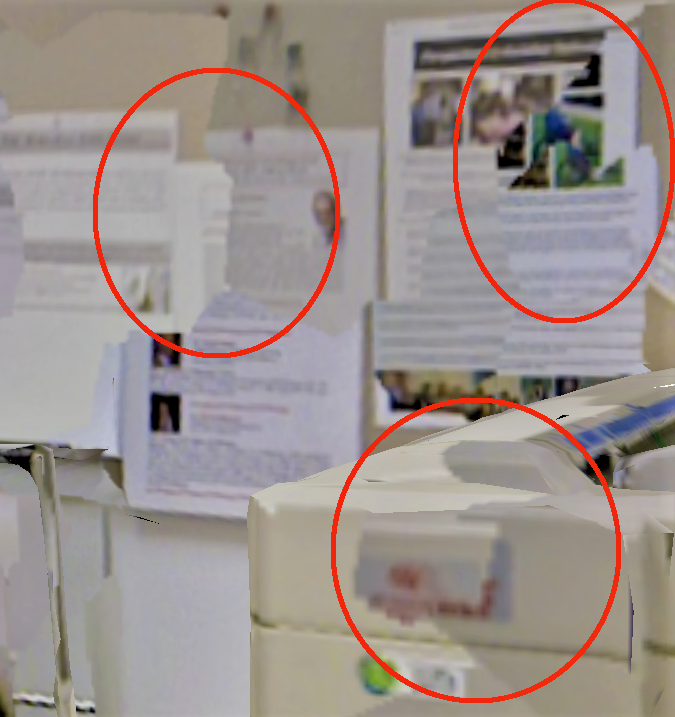}&
\includegraphics[width = 0.45\columnwidth]{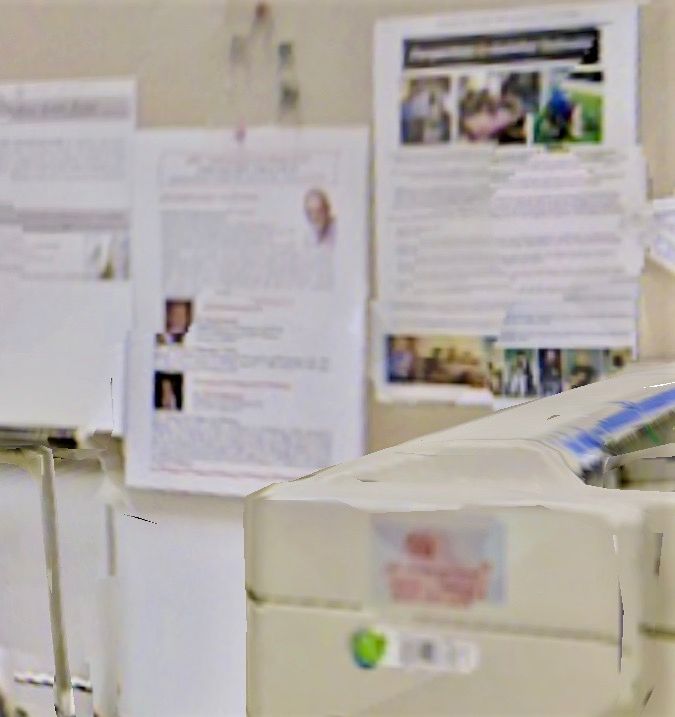}
\end{tabular}
\caption{Texture reconstruction: $(left)$ direct mapping using closest keyframes; $(right)$ our result after alignment and colour correction.}
\label{fig:teaser}
\end{figure}

% problems: colour correction & pose correction
Texture alignment can be seen as an image stitching problem in 3D between several keyframes, where moving any keyframe may impact several neighbours. Hence, an efficient method should find the optimal alignment by considering all keyframes at the same time. Such a method can be very slow and may trap in some local minimum. In this paper we find the global texture assignment using a non-iterative method, which exploits 3D feature matching. Our method has a very low computational complexity as it only requires solving a sparse system of linear equations. Figure~\ref{fig:teaser} shows how visual seams, caused by errors in camera poses, are removed after applying the optimal alignment and colour correction.

%%__________________________________________________________________________________
\section{State of the Art}
\label{sec:soa}

\begin{figure*}%[b]
\centering
\begin{tabular}{cc}
\includegraphics[width=0.35\linewidth]{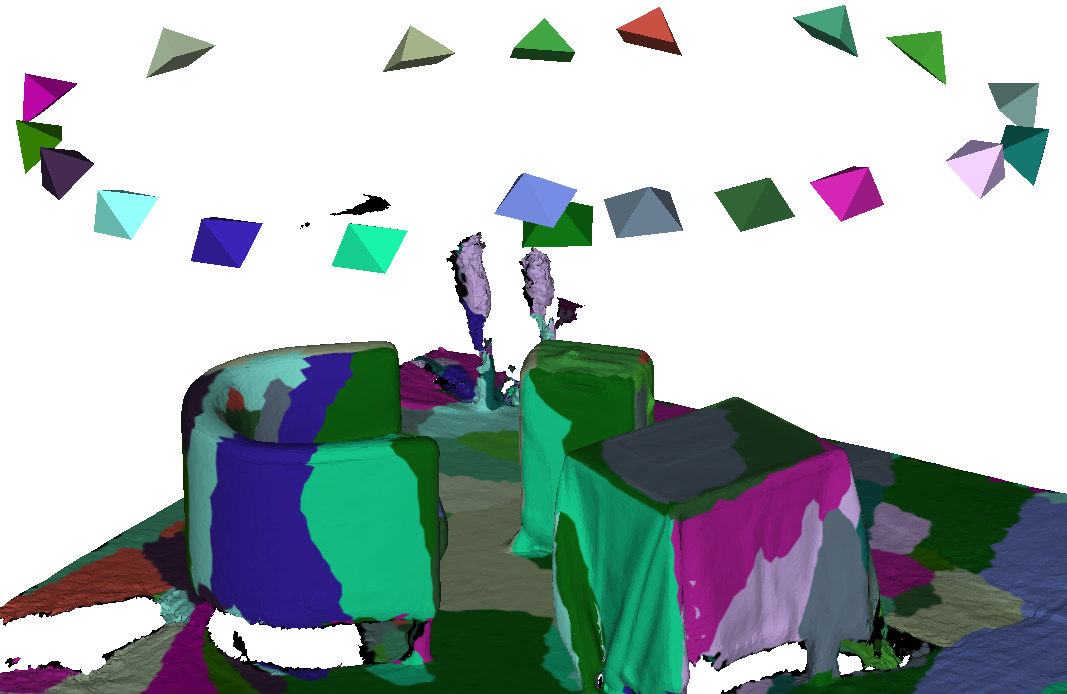}&
\includegraphics[width=0.35\linewidth]{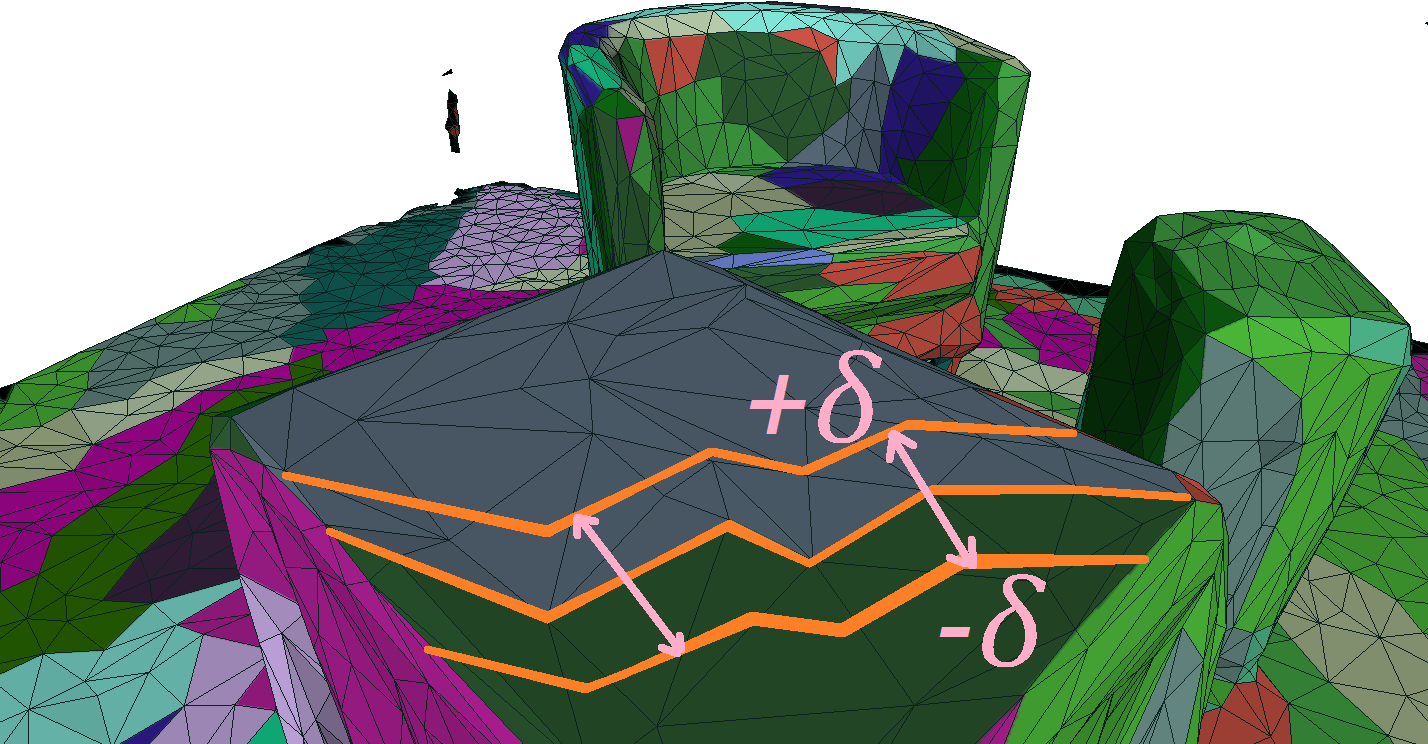}\\
\end{tabular}
\caption{View selection: $(left)$ scene decomposed into fragments using automatic view selection; $(right)$ margin around a fragment border.}
\label{fig:views}
\end{figure*}

Texture reconstruction is widely studied in both computer vision and graphics communities \cite{fu18, kim19}. Given a mesh and a camera trajectory, the first step consists of scoring the texture quality using the camera distance and angle \cite{lempitsky07}. Then, a greedy algorithm may be applied for texturing each region using the closest view. Markov Random Field (MRF), alternatively, proposes an optimal way that reduces seam effects in the texture map \cite{huang17}. It decomposes the scene into small \textit{fragments}, each textured by the best keyframe that is closer and more fronto-parallel (see \autoref{fig:views}).

Illumination changes across the keyframes may lead to visual seams, which requires colour correction~\cite{waechter14, rouhani18}. In addition, there might be geometric misalignment between fragments due to the errors in camera pose estimation and geometry reconstruction~\cite{bi17}. The MRF formulation in~\cite{gal10} not only selects the best view per face, but also corrects the projection within $9$ possible translations. Optical flow between the overlapping images is used in~\cite{dellepiane12} to warp input images and correct local misalignment. 

Zhou et al. in~\cite{zhou14} propose a method to jointly find the optimal camera poses and the best colour per vertex. It requires minimizing a non-linear least squares that converges slowly. Similarly, Fu et al. in ~\cite{fu18} present a two-step method for global and local texture alignment, where deformations per both fragment and vertex are permitted. Laplacian model is used in~\cite{li18} to capture non-rigid deformations of keyframes and to obtain a better texture consistency along the fragment borders. However, both \cite{li18,fu18} are iterative methods that get regularly stuck in local minimum. In the presence of large geometric errors, their correction methods fail to find the optimal alignment and generate some local texture distortions.

In a recent work, Lee et al. in \cite{lee20} propose a real-time texture integration framework for on-line RGB-D scanning. In their fusion algorithm they update the texture map in every step of geometry fusion, which requires updating the camera motion field in a hierarchical manner. The photometric consistency of the texture and input images is evaluated in their texture-to-image registration step; therefore, the quality of texture reconstruction may degrade a lot in the presence of exposure changes and view-dependent appearance.

\section{Proposed Texturing Pipeline}
\label{sec:proposed}

The input of the texture mapping consists of a 3D mesh with lists of vertices $\mathcal{V}$ and faces $\mathcal{F}$ in addition to several keyframes $\{I_1, \dots, I_N\}$ and their poses. The proposed method, summarized in \textbf{Algorithm 1}, starts by extracting 3D feature points: for each keyframe we extract the 2D keypoints using SURF \cite{brown07} and back-project them to 3D space using the \textit{virtual depth map} obtained by rendering the mesh in the given keyframe pose.

\begin{algorithm}%[H]
\begin{algorithmic}[1]
\STATE \textbf{Input:} 3D mesh, keyframes + poses, margin size.
\STATE \textbf{Step 0. 3D keypoint extraction:}
\STATE Extract and back-project keypoints for each keyframe.
\STATE \textbf{Step 1. View selection:}
\STATE Find the best keyframe per face.
\STATE \textbf{Step 2. Texture alignment:}
\STATE \textbf{a. Construct the pose correction system:}
\FOR{every fragment border}
   \STATE i. Select visible keypoints within the margin.
   \STATE ii. Match the selected feature points.
   \STATE iii. Update the correction system for all matching pairs.
\ENDFOR
\STATE \textbf{b. Solve the system of equations} to correct fragment poses.
\STATE \textbf{c. Build the texture atlas} for the corrected poses \& colours.

\end{algorithmic}
\caption{Texture reconstruction with alignment.}
\end{algorithm}

\subsection{View Selection}
\label{subsec:pa1}

In view selection, for each face, we aim at finding the keyframe that will provide the best texturing. Given the labeling $\mathbf{l}=\left(l_f\right)_{f\in\mathcal{F}}$ with $l_f \in \left[1, \dots, N\right]$, we consider the following objective function :

\begin{equation}                                                  
E\left(\mathbf{l}\right)=\sum_{f\in\mathcal{F}}{\psi_f\left(l_f\right)\ }+\lambda_1\sum_{\left(f,g\right)\in\mathcal{E}}{\varphi_{fg}\left(l_f,l_g\right)} \label{eq:MRF},
\end{equation}
where $\mathcal{E}$ is the set of neighbor faces and $\lambda_1$
is a positive parameter that weights the influence of the second term. The data term $\psi_f\left(l_f\right)$ measures the quality of the keyframe $I_{l_f}$ for texturing the face $f$ and the regularization term $\varphi_{fg}\left(l_f,l_g\right)$ controls the texture smoothness between faces $f$ and $g$. In our implementation, we use the area of projection as the data term and the Potts model as the regularizer \cite{rouhani18}. 
We minimize the global energy by applying graph cuts~\cite{boykov01}. The resulting labeling decomposes the mesh into small fragments, a fragment being a set of connected faces with the same label.

\subsection{Texture Alignment System}
\label{subsec:2}

In this section we build an over-determined system of equations $\mathbf{A}_{pose}\boldsymbol\Omega = \mathbf{b}_{pose}$, where $\boldsymbol\Omega$ consists of all correction vectors $\boldsymbol\omega_i$ of each fragment $i$. Every correction vector $\boldsymbol\omega_i = [\alpha, \beta, \gamma,t_x, t_y, t_z]^T$ has $6$ parameters building a skew-symmetric matrix $\mathbf{T_i}$:

\begin{equation}                                                  
\mathbf{T}_i=
\begin{pmatrix}
    1 & -\gamma & \beta & t_x \\
    \gamma & 1 & -\alpha & t_y \\
    -\beta & \alpha & 1 & t_z \\
    0 & 0 & 0 & 1
\end{pmatrix}.
\label{eq:Ti}
\end{equation}

For the sake of simplicity in the notations, let us note abusively $I_i$ the keyframe associated with fragment $i$. For every pair of neighboring fragments $i$ and $j$ we follow these steps to update the matrix $\mathbf{A}_{pose}$ and vector $\mathbf{b}_{pose}$:
\newline

\noindent\textbf{i) Select relevant keypoints}: First, we find all keypoints in $I_i$ and $I_j$ that fall inside a given margin around their common border (see \autoref{fig:views}$(right)$). Moreover, we filter out those keypoints that are not visible in the other keyframe. 
\newline

\begin{figure}%[t]
\centering
\includegraphics[width=.9\columnwidth]{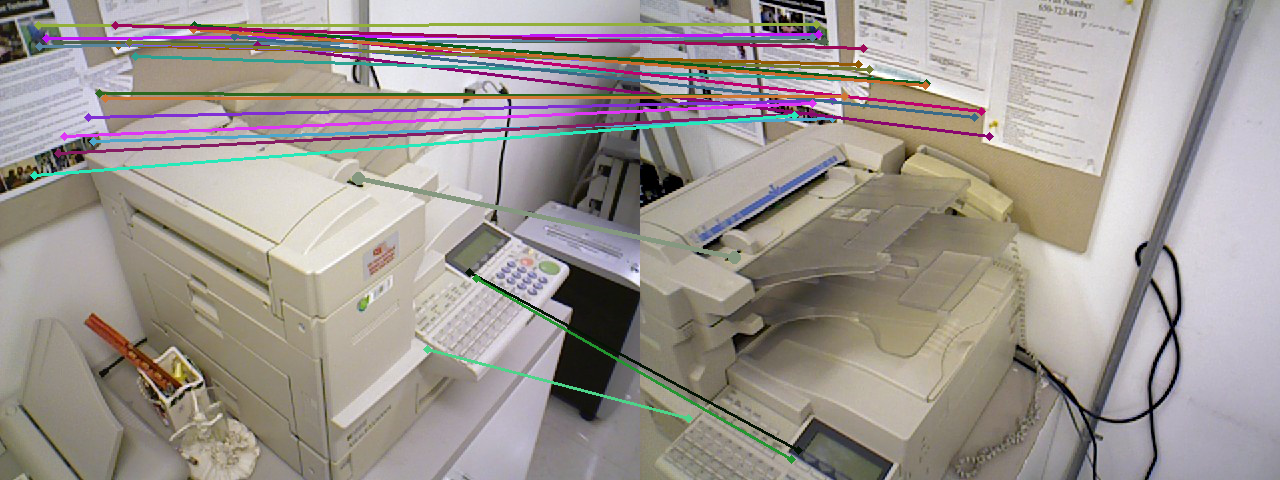}
\caption{Geometry-aware matching around a fragment border.}
\label{fig:matching}
\end{figure}

\noindent\textbf{ii) Match selected keypoints}: These two sets of keypoints are matched using FLANN \cite{muja09}. 3D coordinates of keypoints are exploited as additional descriptors for discarding those far matches. Then, we obtain a set of pairs of 3D keypoints $\{(\mathbf{P}^i_k,\mathbf{P}^j_k)\}$ with weights $\{\mu_k\}$ that are higher close to the border (see \autoref{fig:matching}).
\newline

\noindent\textbf{iii) Update the pose correction system}: These matching pairs must impact $\mathbf{T}_i$ and $\mathbf{T}_j$: the pose correction of the neighboring fragments. Ideally, $\mathbf{P}^i_k$ and $\mathbf{P}^j_k$ must get closer after applying these corrections:
  
\begin{equation}                                                  
\mathbf{P}^i_k+\mathbf{A}^i_k \boldsymbol\omega_i  = \mathbf{P}^j_k +\mathbf{A}^j_k \boldsymbol\omega_j,
\label{eq:Match}
\end{equation}
where $\mathbf{A}^i_k$ is derived from $\mathbf{P}^i_k=(x_k, y_k, z_k)$ as follows: 
 
\begin{equation}                                                  
\mathbf{A}^i_k=
\begin{pmatrix}
    0 & z_k & -y_k & 1 & 0 & 0\\
    -z_k & 0 & x_k & 0 & 1 & 0\\
    y_k & -x_k & 0 & 0 & 0 & 1
\end{pmatrix}.
\label{eq:Ai}
\end{equation}
Then, every pair of matches adds $3$ constraints by copying $\mu_k\mathbf{A}^i_k$ and $-\mu_k\mathbf{A}^j_k$ to the associated columns of $\mathbf{A}_{pose}$ and by concatenating the vector $\mu_k(\mathbf{P}^j_k-\mathbf{P}^i_k)$ to $\mathbf{b}_{pose}$.

Having considered all pairs of neighbouring fragments, the optimal alignment can be found by minimizing: 
\begin{equation}                                                  
\varphi({\boldsymbol\Omega}) = \| \mathbf{A}_{pose}\boldsymbol\Omega -\mathbf{b}_{pose} \|^2 +\lambda_2 \|\boldsymbol\Omega\|^2,
\label{eq:LS}
\end{equation}
where the regularization factor $\lambda_2$ controls the amount of correction. It leads to a sparse system of equations that can be quickly solved by QR factorization.

The optimal correction vector $\boldsymbol\Omega$ encodes the deformation matrices $\mathbf{T}_i$ for each fragment.
Moreover, we apply colour correction \cite{rouhani18} before copying the texture information from each keyframe to the final texture atlas (see \autoref{fig:Lounge}).

\section{Experimental Results}
\label{sec:exper}

\begin{figure}%[b]
\centering
\includegraphics[width=0.48\columnwidth]{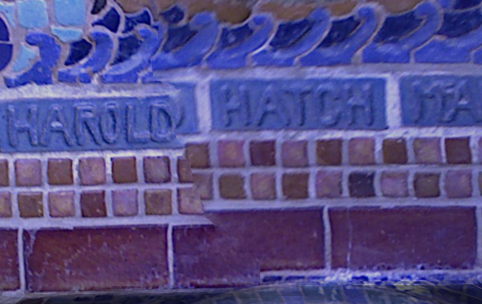}
\includegraphics[width=0.48\columnwidth]{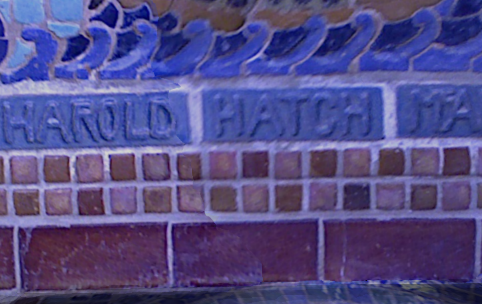}
\caption{$(left)$ Iterative method, such as \cite{li18}, may get trapped in a local minimum, $(right)$ while our method finds the global solution.}
\label{fig:result1}
\end{figure}

\begin{figure}%[b]
\centering
\includegraphics[width=0.48\columnwidth]{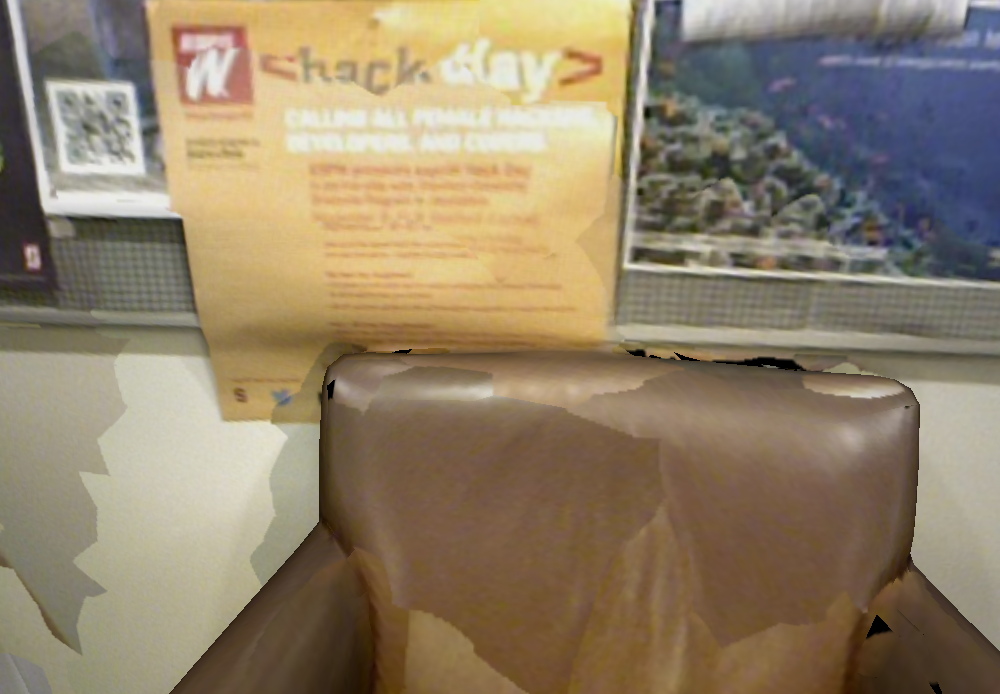}
\includegraphics[width=0.48\columnwidth]{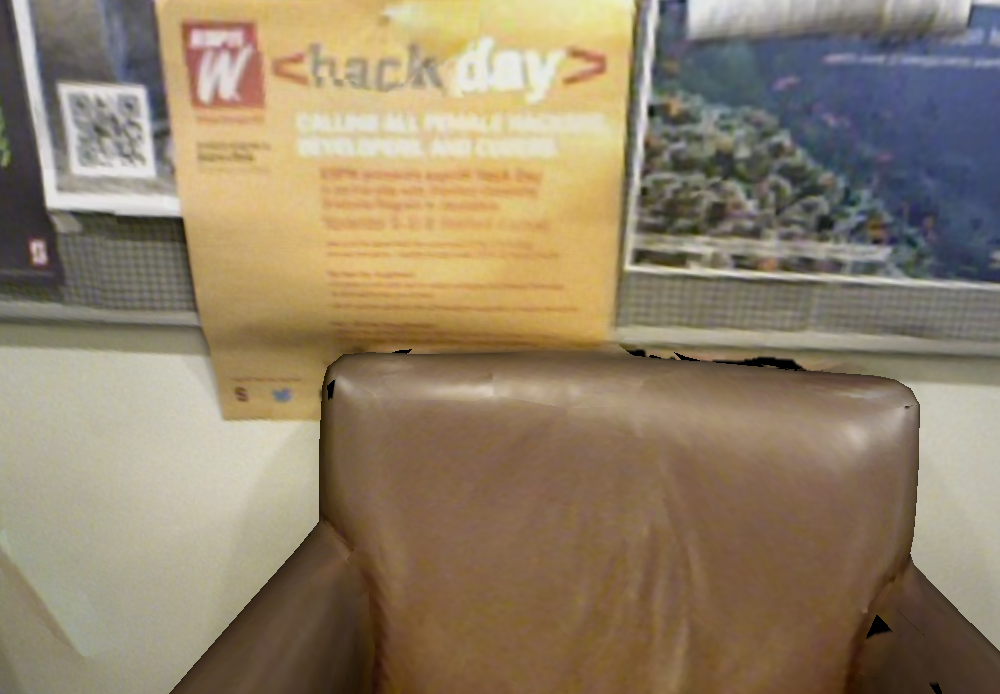}
\caption{$(left)$ Input; $(right)$ colour correction and alignment applied.}
\label{fig:Lounge}
\end{figure}

\begin{figure*}%[t]
\centering
\begin{tabular}{cccc}
\hspace{-8pt}\includegraphics[width=.24\linewidth]{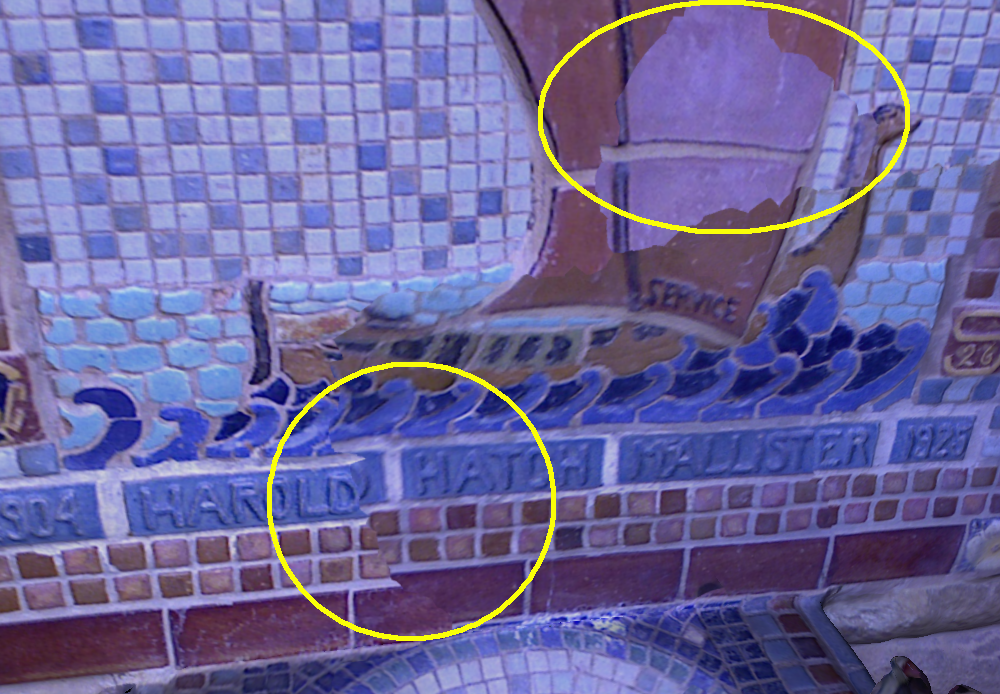}&
\hspace{-8pt}\includegraphics[width=.24\linewidth]{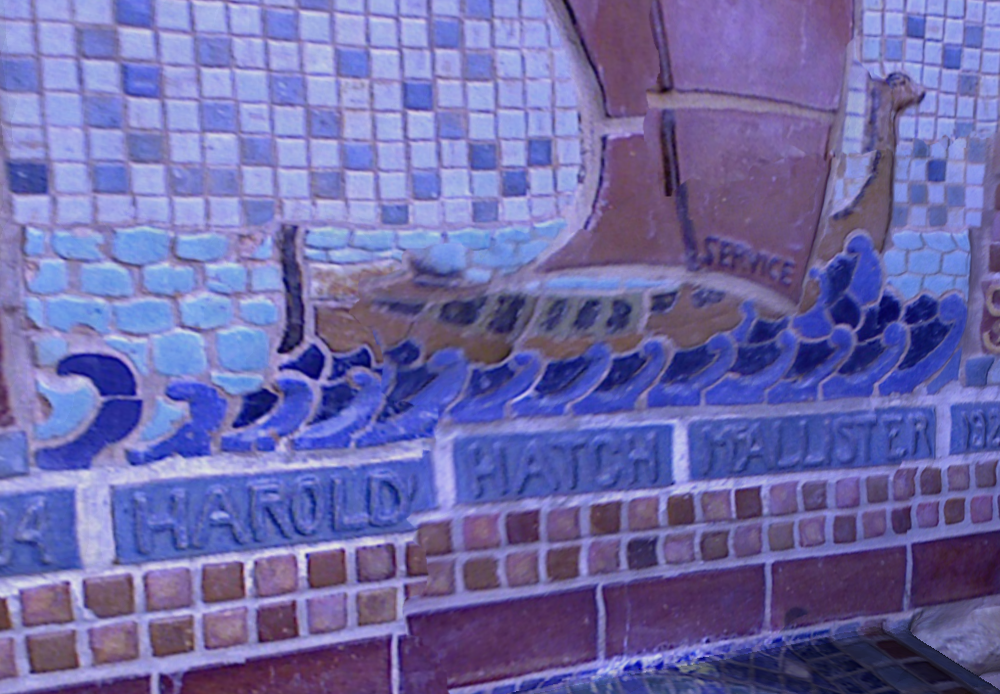}&
\hspace{-8pt}\includegraphics[width=.24\linewidth]{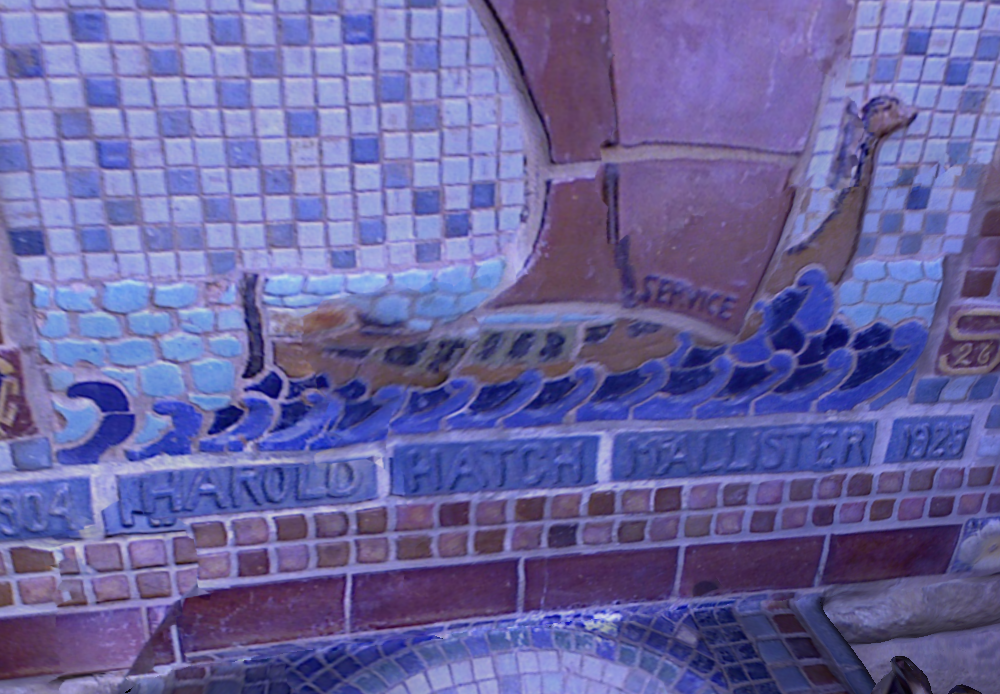}& 
\hspace{-8pt}\includegraphics[width=.24\linewidth]{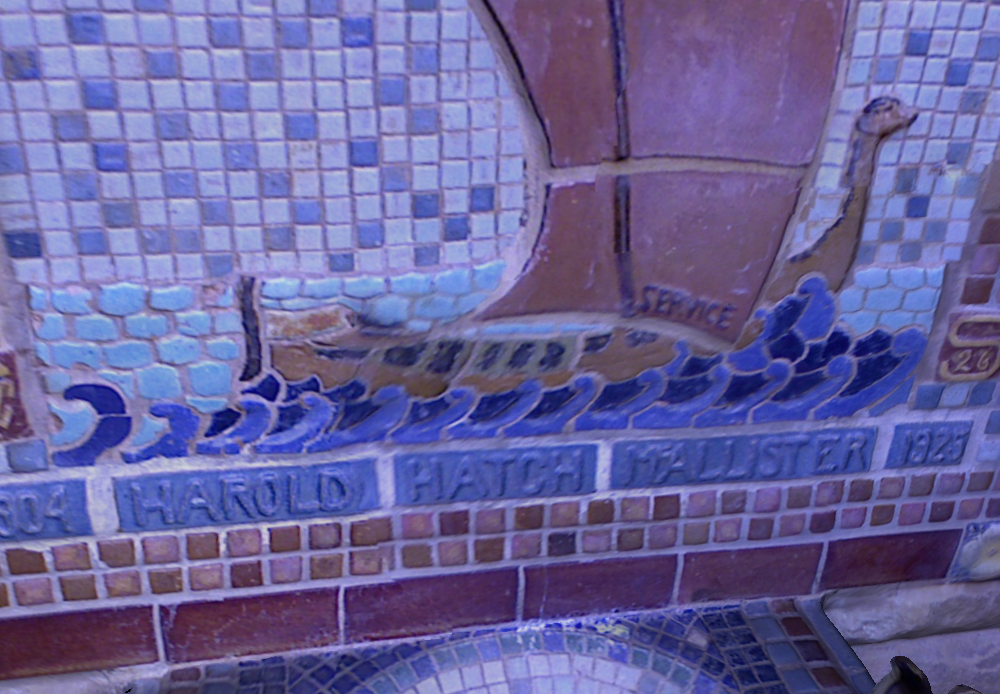}\\

\hspace{-8pt}\includegraphics[width=.24\linewidth]{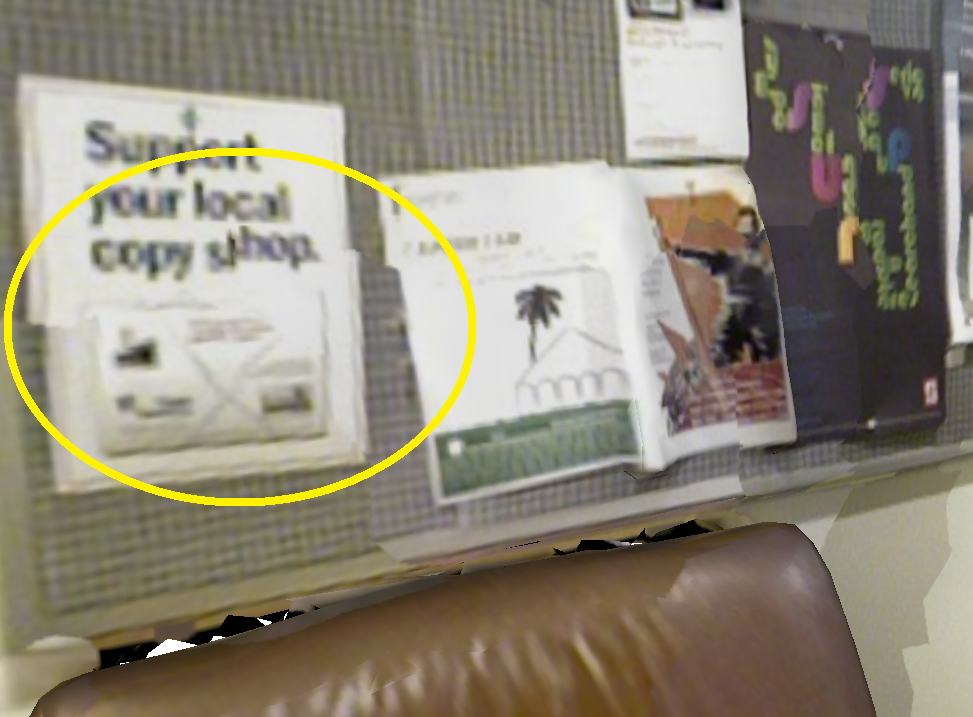}&
\hspace{-8pt}\includegraphics[width=.24\linewidth]{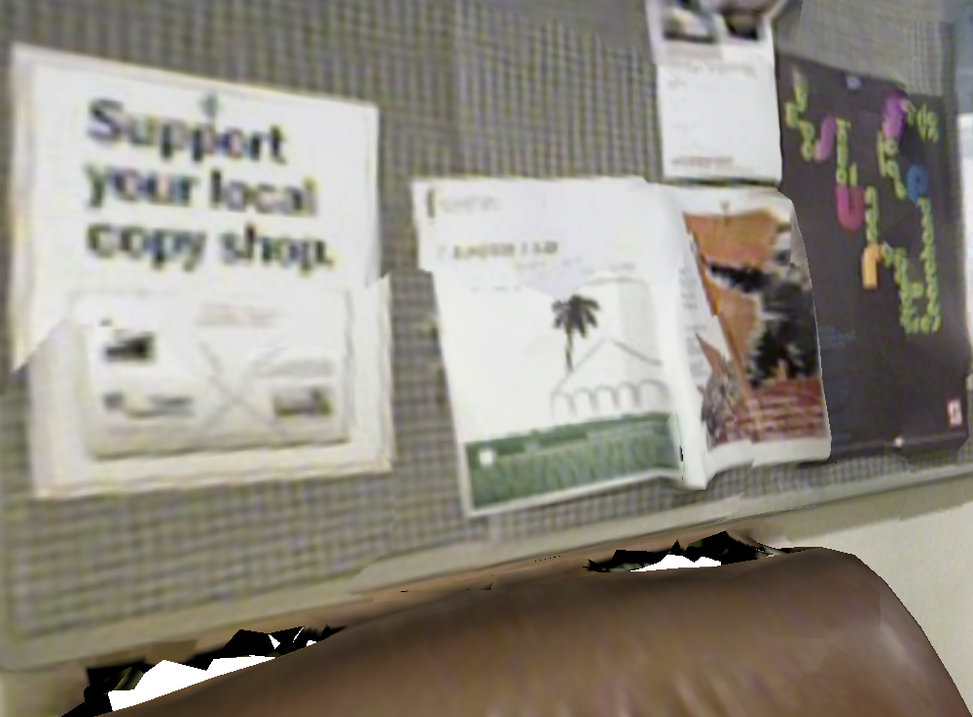}&
\hspace{-8pt}\includegraphics[width=.24\linewidth]{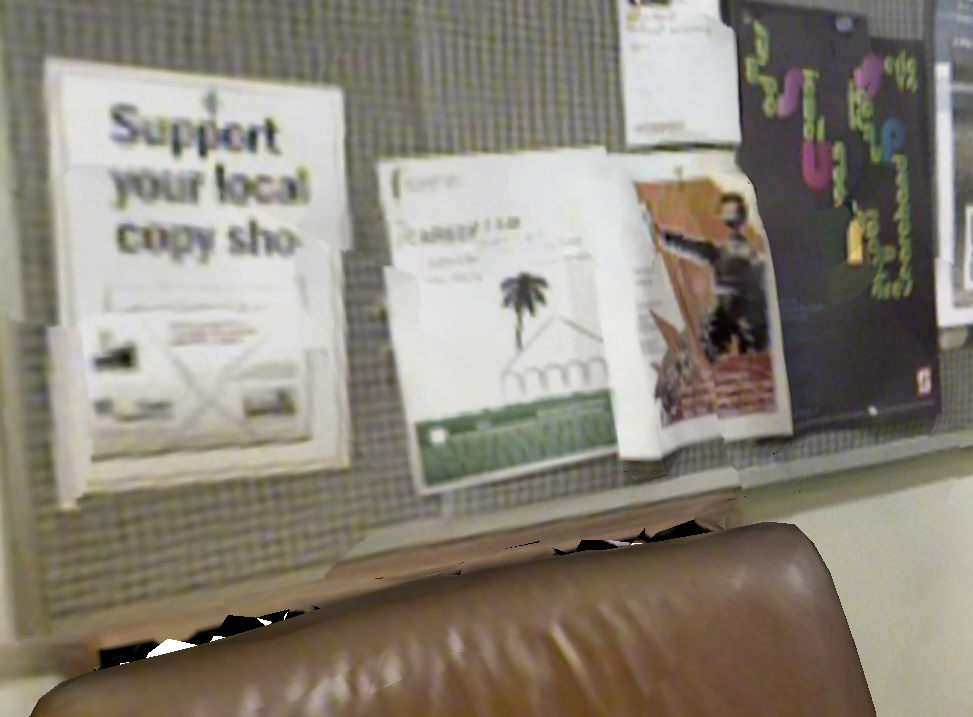}&
\hspace{-8pt}\includegraphics[width=.24\linewidth]{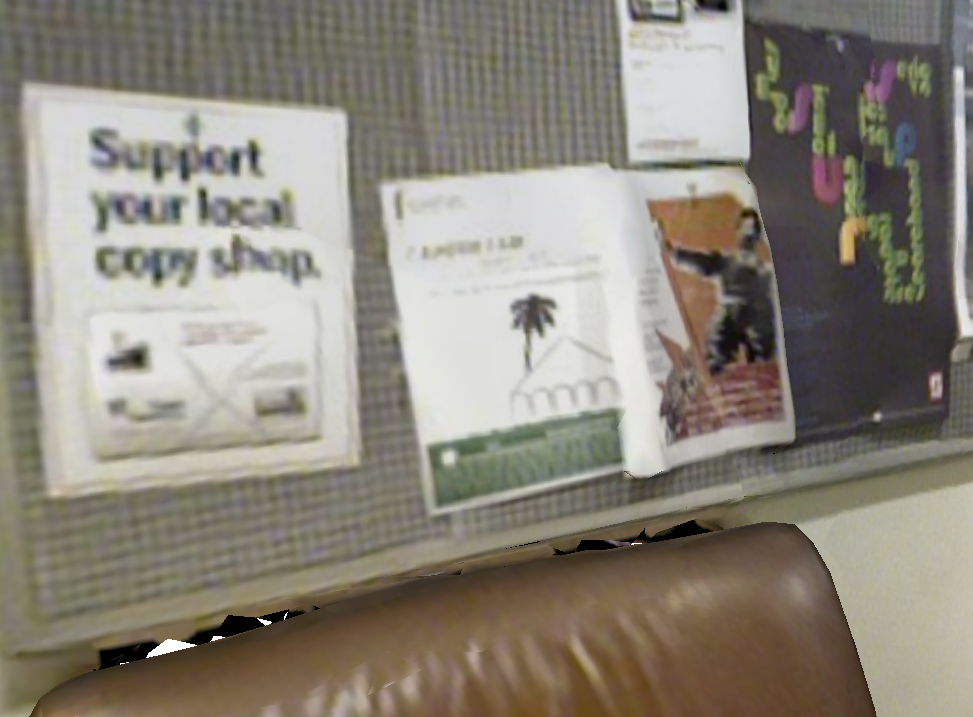}\\

\hspace{-8pt}\includegraphics[width=.24\linewidth]{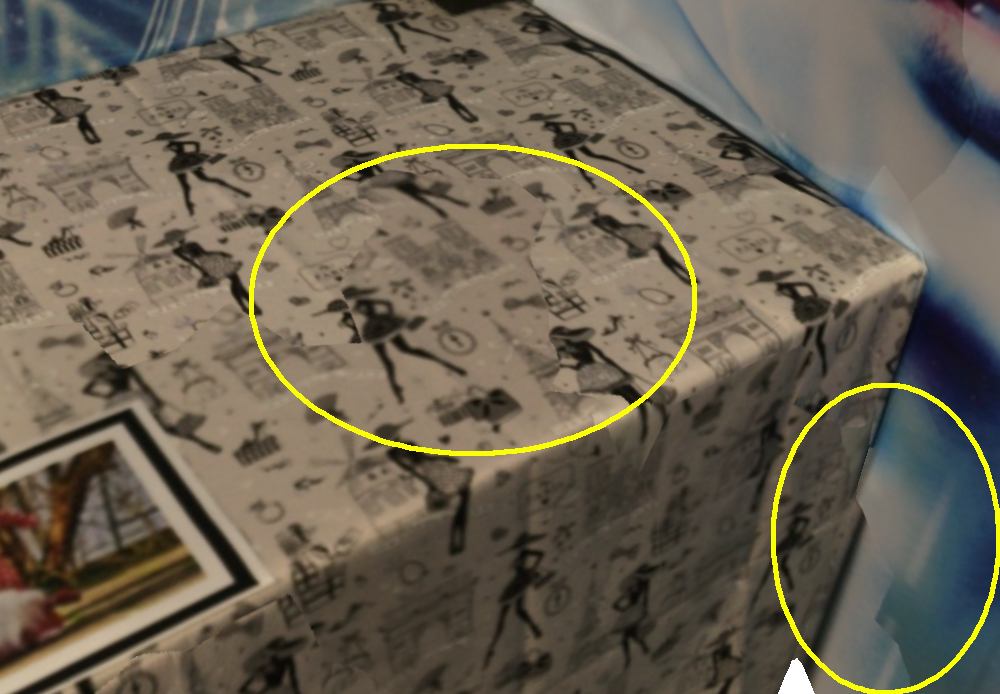}&
\hspace{-8pt}\includegraphics[width=.24\linewidth]{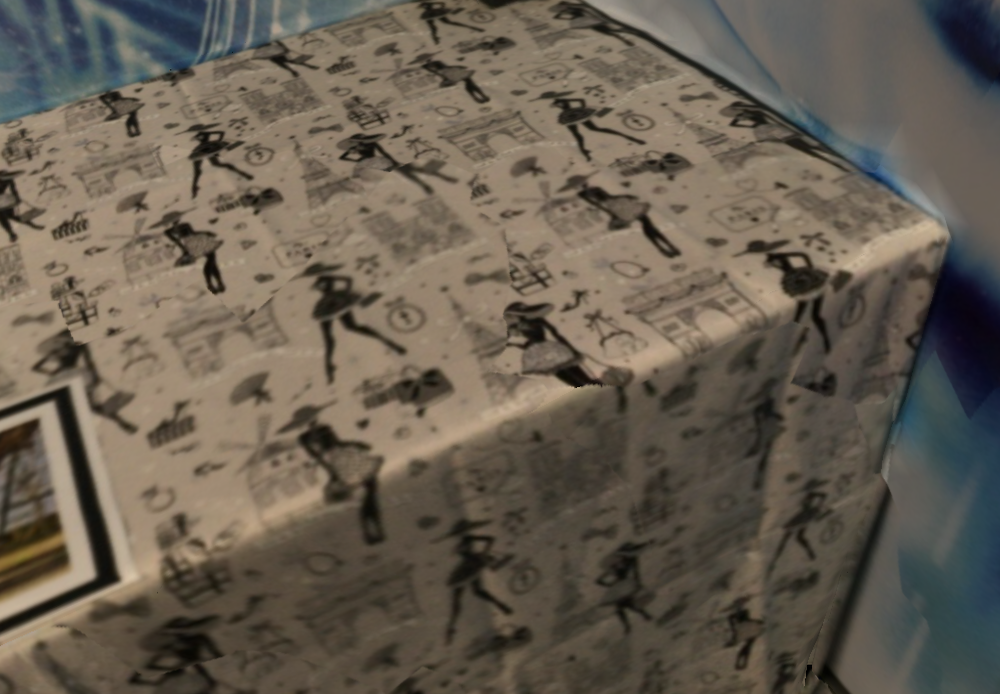}&
\hspace{-8pt}\includegraphics[width=.24\linewidth]{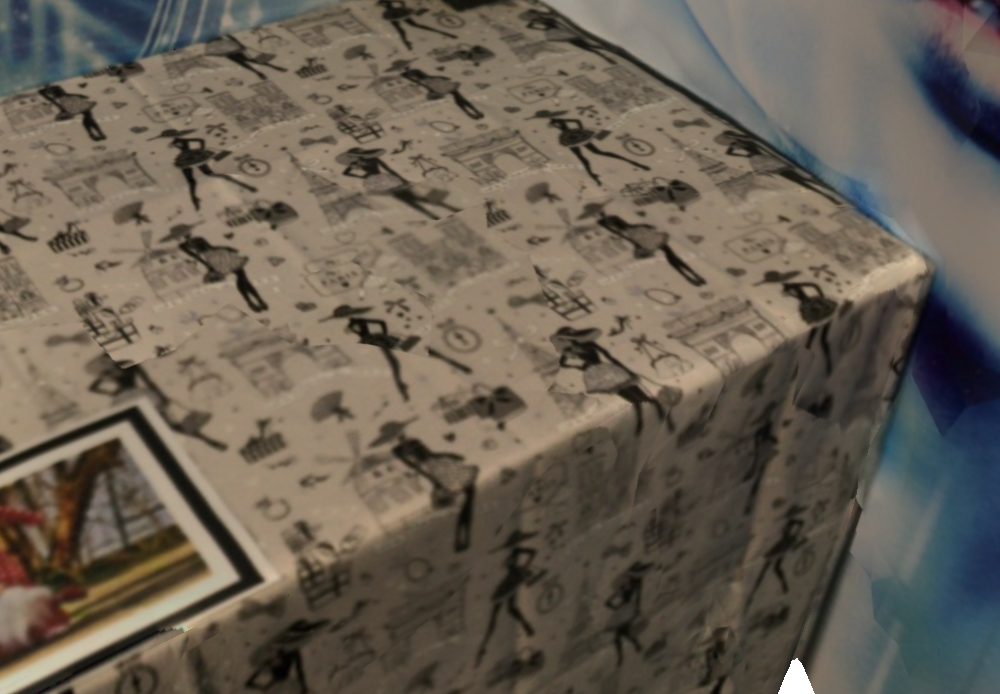}& 
\hspace{-8pt}\includegraphics[width=.24\linewidth]{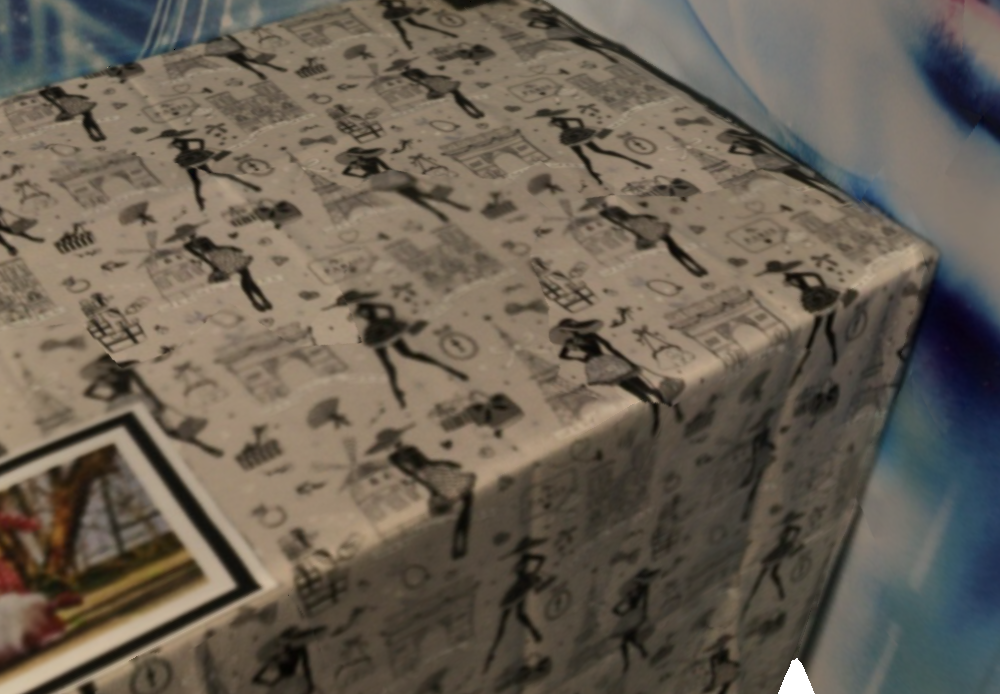}\\

\hspace{-8pt}\includegraphics[width=.24\linewidth]{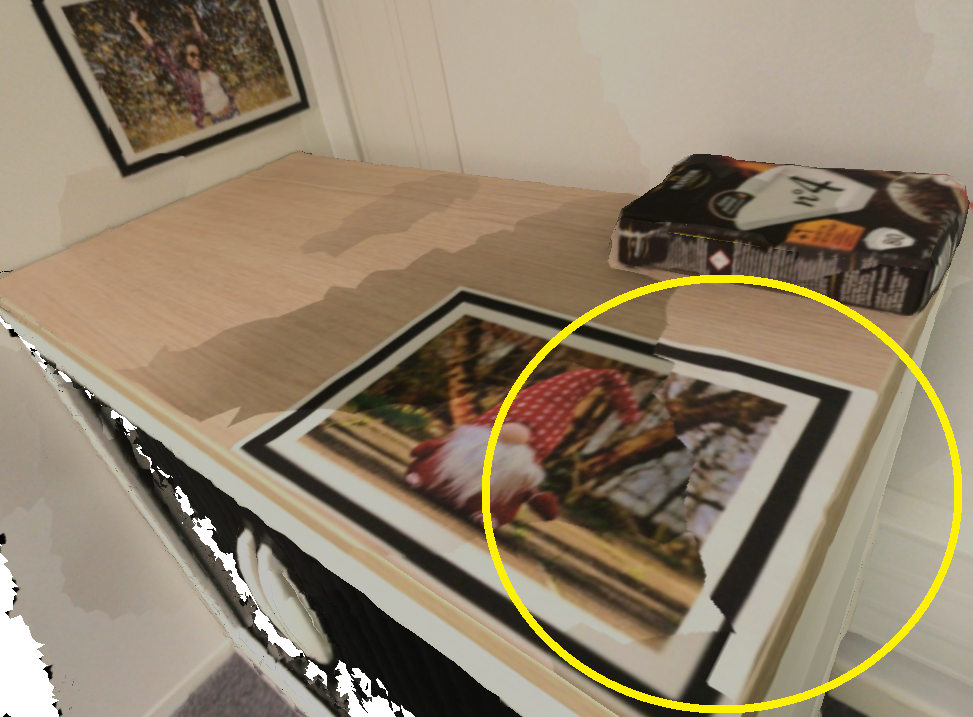}&
\hspace{-8pt}\includegraphics[width=.24\linewidth]{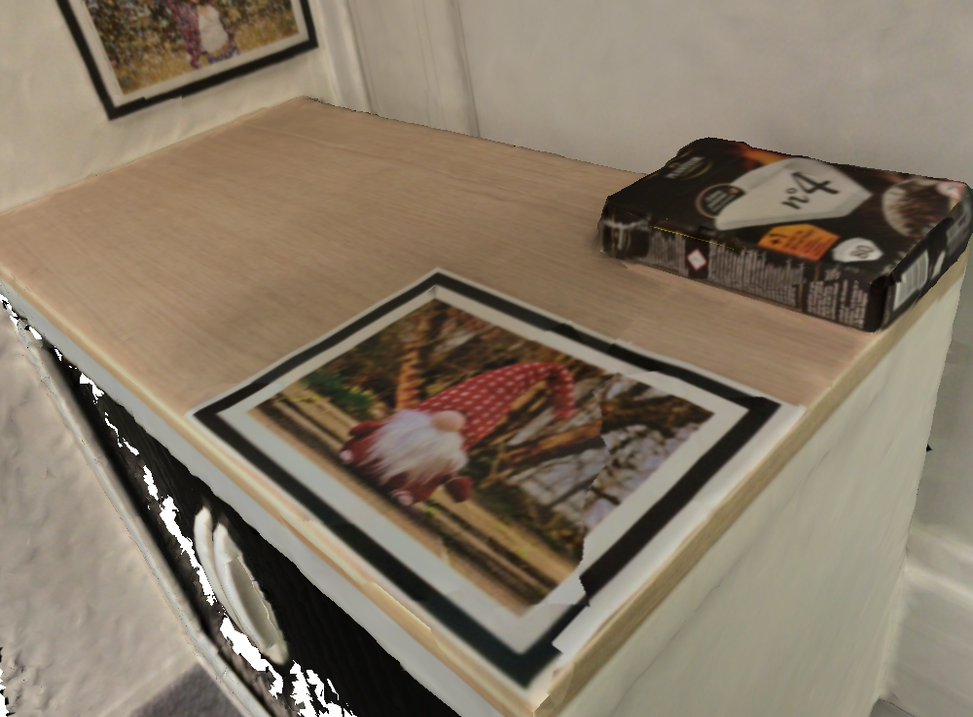}&
\hspace{-8pt}\includegraphics[width=.24\linewidth]{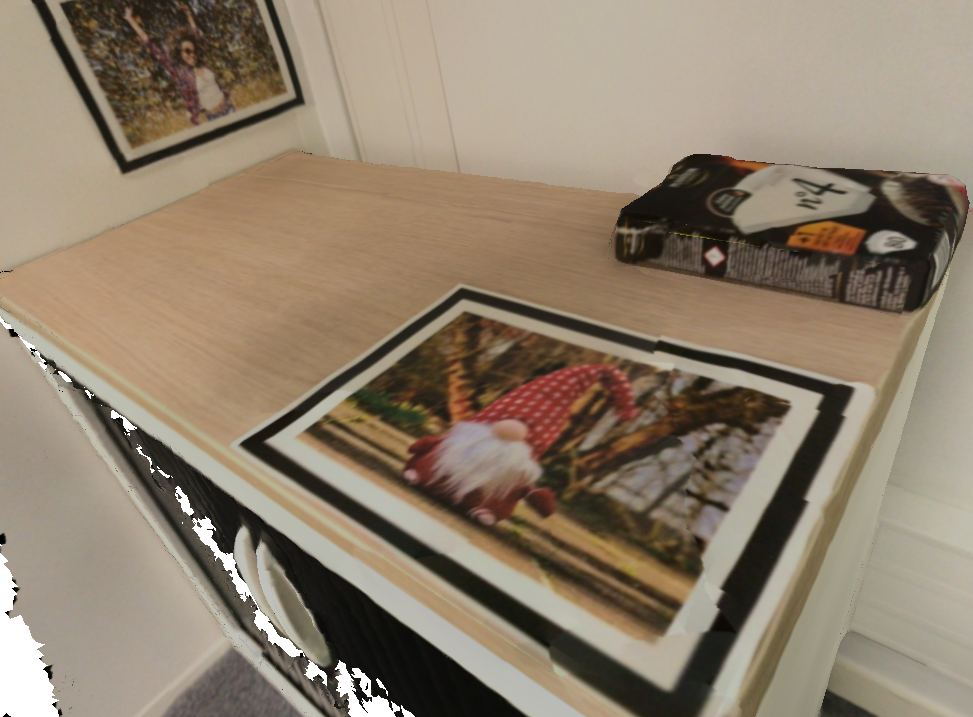}& 
\hspace{-8pt}\includegraphics[width=.24\linewidth]{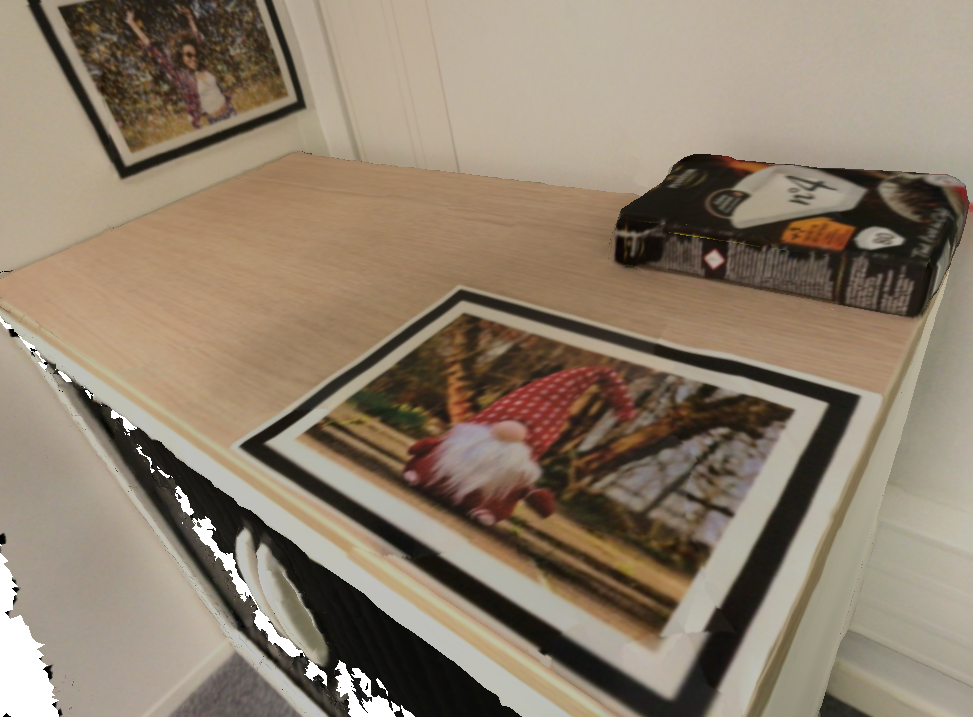}\\

\hspace{-8pt}\includegraphics[width=.24\linewidth]{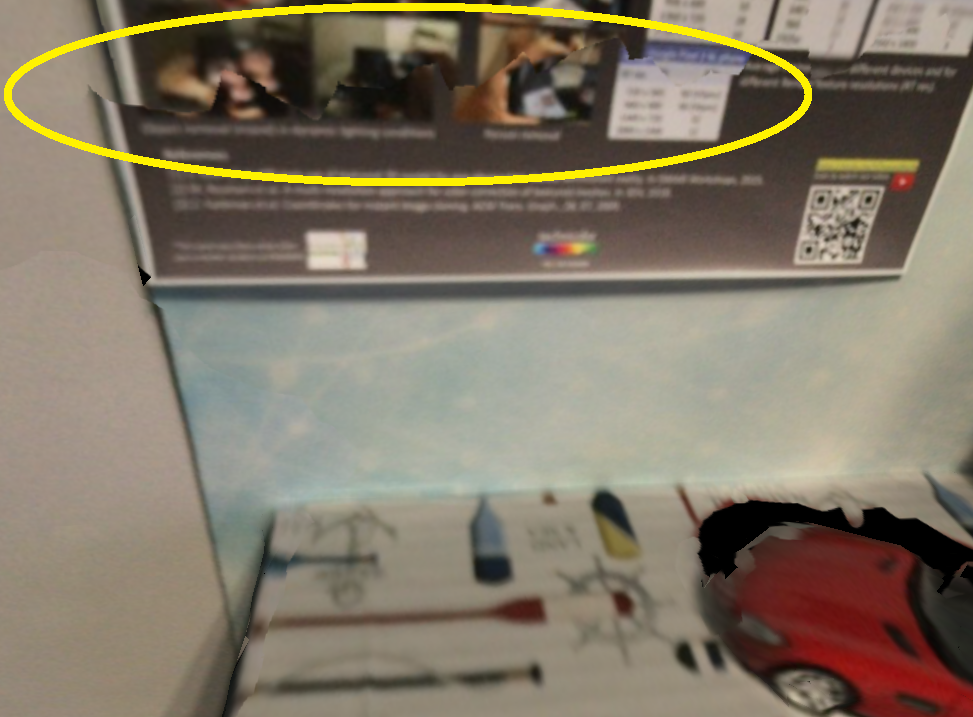}&
\hspace{-8pt}\includegraphics[width=.24\linewidth]{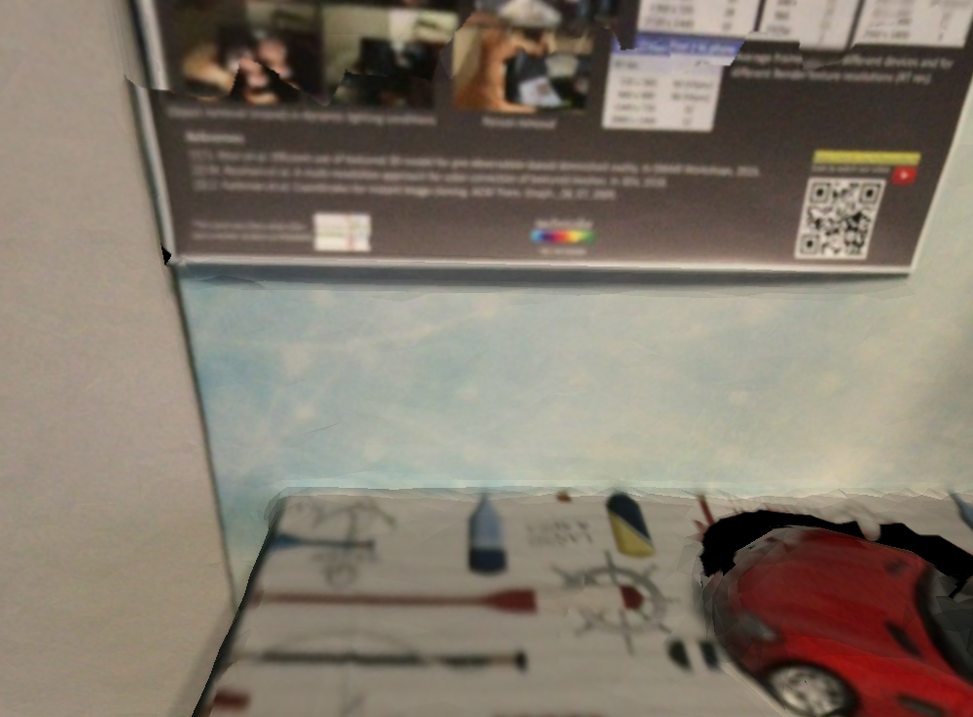}&
\hspace{-8pt}\includegraphics[width=.24\linewidth]{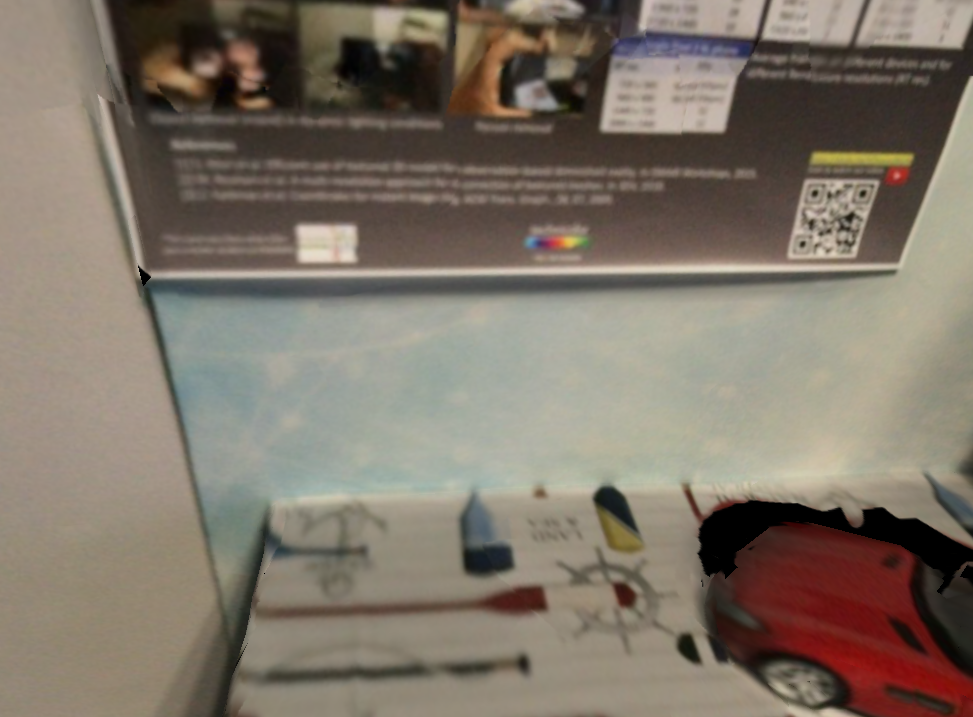}& 
\hspace{-8pt}\includegraphics[width=.24\linewidth]{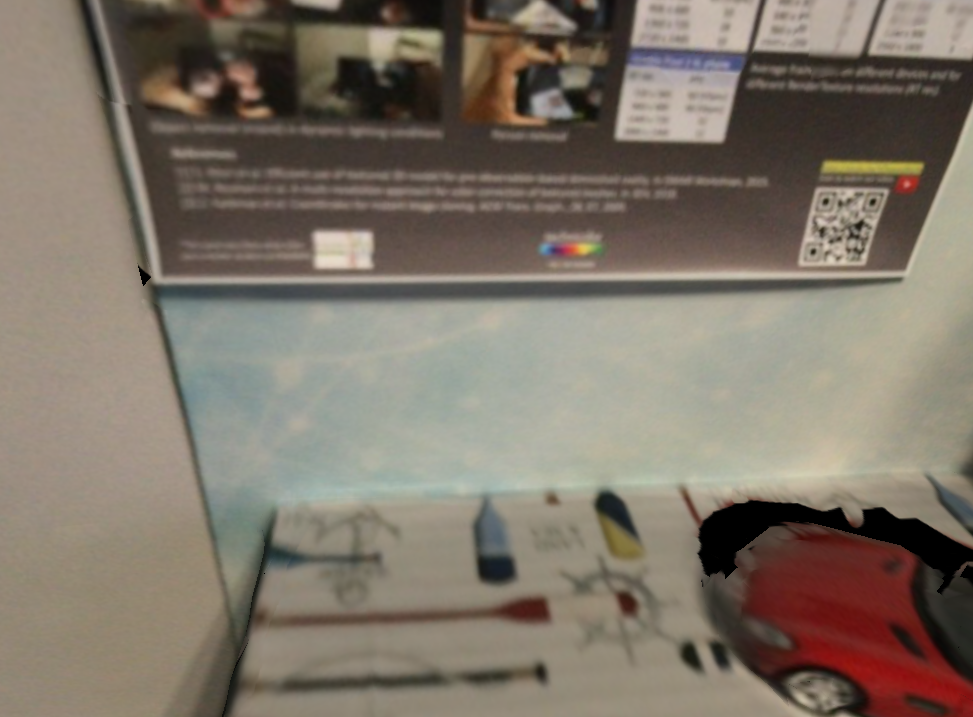}\\

\hspace{-8pt}\includegraphics[width=.24\linewidth]{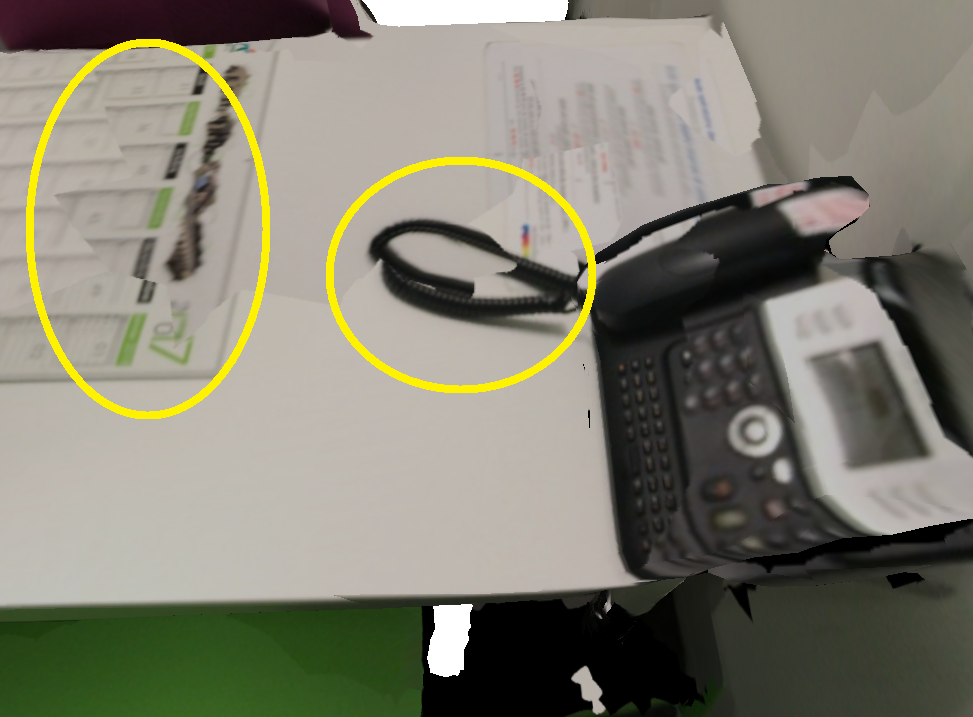}&
\hspace{-8pt}\includegraphics[width=.24\linewidth]{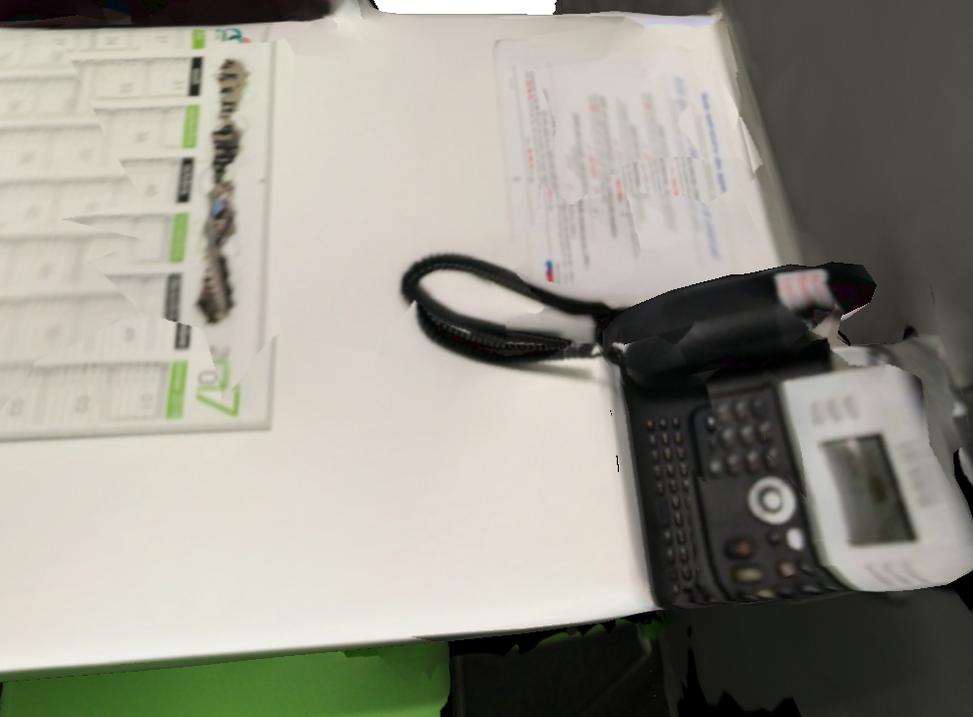}&
\hspace{-8pt}\includegraphics[width=.24\linewidth]{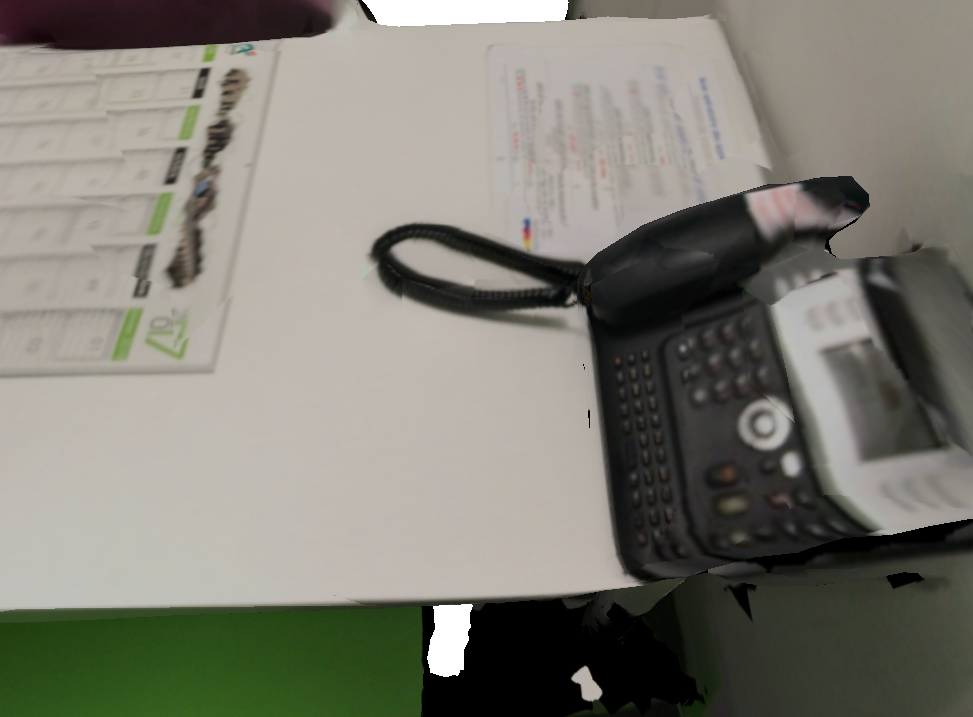}& 
\hspace{-8pt}\includegraphics[width=.24\linewidth]{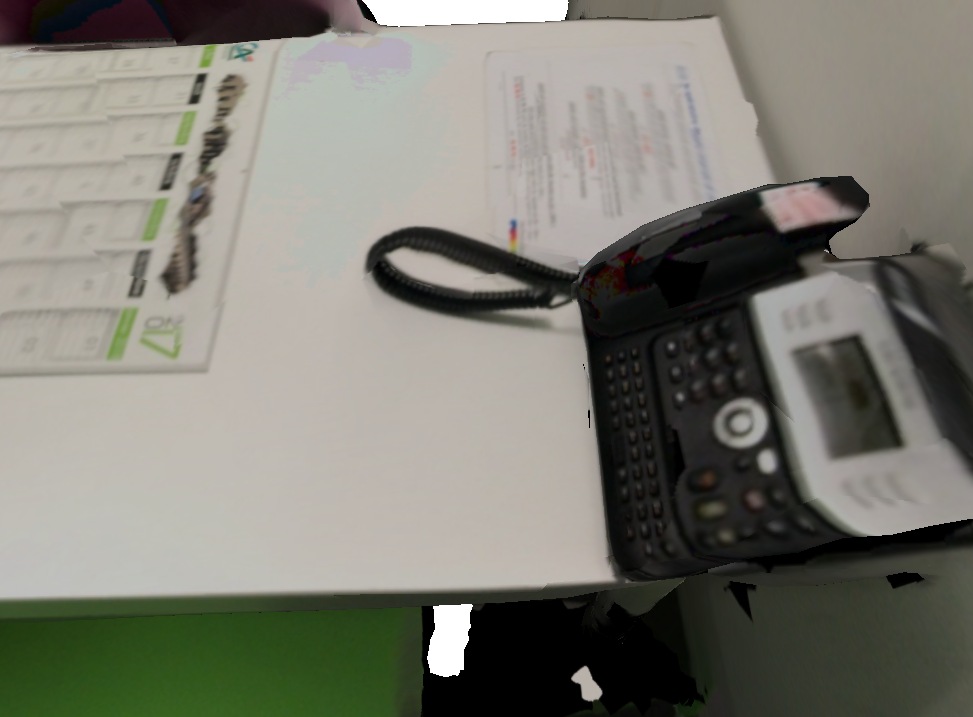}\\
(a)&(b)&(c)&(d)
\end{tabular}

\caption{Comparisons: $(a)$ direct texture mapping before the colour correction \cite{waechter14}; $(b)$ method in \cite{allene08} encouraging colour consistency; $(c)$ method in \cite{gal10} applying pose correction per face; $(d)$ our results after pose and colour correction.}
\label{fig:result3}
\end{figure*}

The proposed method has been tested on several public data sets, where a raw mesh and multiple keyframes are provided. Figures \ref{fig:teaser}, \ref{fig:result1}, \ref{fig:Lounge} and the first two rows of \autoref{fig:result3} illustrate the texturing results for \textit{Copyroom}, \textit{Fountain} and \textit{Lounge} provided by \cite{zhou14}. In addition, we gathered our own data sets using a StructureIO mounted on an iPad Pro. The last four rows of \autoref{fig:result3} are captured through our scanning platform.

The amount of distortion in Figures \ref{fig:teaser} and \ref{fig:result1} makes most pixel-wise or iterative methods, such as \cite{fu18} and \cite{li18}, get trapped in a local minimum. In contrast, our global optimization finds the optimal alignment over the whole set of texture fragments in a non-iterative way by solving a sparse system of linear equations.

Figure~\ref{fig:result3} presents comparisons of the method with several state of the art techniques. The results have been qualitatively compared with \cite{allene08,gal10,waechter14}. The far left column highlights the regions where misalignment and illumination changes happen. In most of cases our method provides the best quality reconstruction. Table \ref{table:cpu} states the details about the scene geometry as well as the CPU timing for different steps. It includes the two main steps of our algorithm for view selection and texture alignment, where global optimization problems are solved by MRF and LS.

Please notice that if timing for our unoptimized step 0 may appear quite long, it is directly related to the chosen type of features to be extracted and could be accelerated preferring ORB to SURF for example. Since we use geometric clues, such as 3D positions of features, for matching the performance does not change by switching between descriptors. Note also that this pre-processing step is highly parallelizable and would deserve a GPU implementation. This implementation change is still an on-going work so that total timings should be very carefully compared.

\begin{table}[h]
\begin{tabular}{l*{7}{c}} 
\toprule

\hspace{-5pt}Fig. & \hspace{-4pt}$|\mathcal{F}|$ & \hspace{-3pt}$N$ & \cite{allene08} & \cite{gal10} & unoptim. & our & our \\
 & & & & & step0 & & step2\\

\hline

\hspace{-5pt}\textbf{ \ref{fig:result3}-$1$} & \hspace{-4pt}117$k$ & \hspace{-3pt}30 & 19.69 & 196.17 & 68.81 & 3.28  & 4.08 \\ 

\hspace{-5pt}\textbf{ \ref{fig:result3}-$2$} & \hspace{-4pt}128$k$ & \hspace{-3pt}55 & 17.11 & 206.09 & 14.74 & 10.06 & 0.81 \\ 

\hspace{-5pt}\textbf{ \ref{fig:result3}-$3$} & \hspace{-4pt}20$k$ & \hspace{-3pt}11  & 0.52 & 7.87 & 20.03 & 0.10 & 0.79 \\ 

\hspace{-5pt}\textbf{ \ref{fig:result3}-$4$} & \hspace{-4pt}81$k$ & \hspace{-3pt}13 & 3.72 & 41.55  & 11.72 & 0.89 & 0.53 \\ 

\hspace{-5pt}\textbf{ \ref{fig:result3}-$5$} & \hspace{-4pt}105$k$  & \hspace{-3pt}7 & 1.84 & 27.55 & 11.09 & 0.71 & 0.54 \\ 

\hspace{-5pt}\textbf{ \ref{fig:result3}-$6$} & \hspace{-4pt}87$k$  & \hspace{-3pt}11 & 2.23 & 23.8  & 10.57 & 0.77 & 0.39 \\ 

\bottomrule 
\end{tabular}
\caption{CPU timing (in seconds) for steps 0-2 of Algorithm 1, and for the methods presented in \cite{allene08} and \cite{gal10} as well. $|\mathcal{F}|$ and $N$ represent the number of faces and keyframes, respectively.}\label{table:cpu}
\end{table}

\section{Conclusion}
\label{sec:conc} 

In this work a novel technique for global texture reconstruction has been presented. Our method is non-iterative and it exploits a geometry-aware feature matching between the keyframes. Moreover, it has a very low computational complexity as it requires solving a sparse system of equations. The margin size around the fragment borders defines how big the overlapping region can be. The qualitative results together with the low computational complexity, and the explicit hyper-parameters prove our method to be efficient and quite fast for texture reconstruction.

\bibliographystyle{abbrv-doi-hyperref-narrow}

\bibliography{refs}

\begin{thebibliography}{10}
\renewcommand*{\sfdefault}{PTSansNarrow-TLF}

\bibitem{allene08}
C.~All{\`e}ne, J.-P. Pons, and R.~Keriven.
\newblock Seamless image-based texture atlases using multi-band blending.
\newblock In {\em Int. Conf. on Pattern Recognition}, pp. 1--4, 2008.

\bibitem{bi17}
S.~Bi, N.~K. Kalantari, and R.~Ramamoorthi.
\newblock Patch-based optimization for image-based texture mapping.
\newblock {\em ACM Trans. on Graphics}, 36(4), 2017.

\bibitem{boykov01}
Y.~Boykov, O.~Veksler, and R.~Zabih.
\newblock Fast approximate energy minimization via graph cuts.
\newblock {\em IEEE Transactions on pattern analysis and machine intelligence},
  23(11):1222--1239, 2001.

\bibitem{brown07}
M.~Brown and D.~G. Lowe.
\newblock Automatic panoramic image stitching using invariant features.
\newblock {\em Int. Jour. Computer Vision}, 74(1):59--73, 2007.

\bibitem{dellepiane12}
M.~Dellepiane, R.~Marroquim, M.~Callieri, P.~Cignoni, and R.~Scopigno.
\newblock Flow-based local optimization for image-to-geometry projection.
\newblock {\em IEEE Trans. Visualization Computer Graphics}, 18(3):463--474,
  2012.

\bibitem{fu18}
Y.~Fu, Q.~Yan, L.~Yang, J.~Liao, and C.~Xiao.
\newblock Texture mapping for {3D} reconstruction with {RGB-D} sensor.
\newblock In {\em Conf. Computer Vision Pattern Recognition}, 2018.

\bibitem{gal10}
R.~Gal, Y.~Wexler, E.~Ofek, H.~Hoppe, and D.~Cohen-Or.
\newblock Seamless montage for texturing models.
\newblock In {\em Computer Graphics Forum}, pp. 479--486, 2010.

\bibitem{huang17}
J.~Huang, A.~Dai, L.~Guibas, and M.~Nie{\ss}ner.
\newblock 3{DL}ite: towards commodity {3D} scanning for content creation.
\newblock {\em ACM Trans. on Graphics}, 2017.

\bibitem{kim19}
J.~Kim, H.~Kim, J.~Park, and S.~Lee.
\newblock Global texture mapping for dynamic objects.
\newblock In {\em Computer Graphics Forum}, pp. 697--705, 2019.

\bibitem{lee20}
J.~H. Lee, H.~Ha, Y.~Dong, X.~Tong, and M.~H. Kim.
\newblock Texturefusion: High-quality texture acquisition for real-time rgb-d
  scanning.
\newblock In {\em IEEE Conf. Computer Vision Pattern Recognition}, pp.
  1272--1280, 2020.

\bibitem{lempitsky07}
V.~Lempitsky and D.~Ivanov.
\newblock Seamless mosaicing of image-based texture maps.
\newblock In {\em IEEE Conf. on Computer Vision and Pattern Recognition}, pp.
  1--6, 2007.

\bibitem{li18}
W.~Li, G.~Huajun, and Y.~Ruigang.
\newblock Fast texture mapping adjustment via local/global optimization.
\newblock {\em IEEE Trans. on Visualization and Computer Graphics}, 2018.

\bibitem{muja09}
M.~Muja and D.~G. Lowe.
\newblock Fast approximate nearest neighbors with automatic algorithm
  configuration.
\newblock In A.~Ranchordas and H.~Ara{\'{u}}jo, eds., {\em VISAPP}, pp.
  331--340. {INSTICC} Press, 2009.

\bibitem{Glen18}
G.~Queguiner, M.~Fradet, and M.~Rouhani.
\newblock Towards mobile diminished reality.
\newblock In {\em {IEEE} Int. Symposium on Mixed and Augmented Reality, {ISMAR}
  2018 Adjunct, Munich, Germany}, pp. 226--231, 2018.

\bibitem{rouhani18}
M.~Rouhani, M.~Fradet, and C.~Baillard.
\newblock A multi-resolution approach for color correction of textured meshes.
\newblock In {\em IEEE Conference on 3D Vision.}, pp. 71--78, 2018.

\bibitem{waechter14}
M.~Waechter, N.~Moehrle, and M.~Goesele.
\newblock Let there be color! large-scale texturing of {3D} reconstructions.
\newblock In {\em European Conference on Computer Vision}, pp. 836--850, 2014.

\bibitem{zhang16}
E.~Zhang, M.~F. Cohen, and B.~Curless.
\newblock Emptying, refurnishing, and relighting indoor spaces.
\newblock {\em ACM Trans. on Graphics}, 35(6):174, 2016.

\bibitem{zhou14}
Q.-Y. Zhou and V.~Koltun.
\newblock Color map optimization for {3D} reconstruction with consumer depth
  cameras.
\newblock {\em ACM Transactions on Graphics}, 33(4):155, 2014.

\end{thebibliography}

\end{document}